\documentclass[conference]{IEEEtran}
\usepackage{booktabs}
\usepackage{xcolor}

\usepackage{cite}

\ifCLASSINFOpdf
  \usepackage[pdftex]{graphicx}
\else
\fi
\usepackage{amssymb}
\usepackage{amsmath}
\usepackage[ruled,vlined]{algorithm2e}
\usepackage{amsmath}
\usepackage{algpseudocode}
\usepackage{subcaption}
\usepackage{dblfloatfix}

\usepackage{hyperref}
\hypersetup{linkcolor=black,urlcolor=blue}

\usepackage{cleveref}
\usepackage{multirow}
\usepackage{diagbox}
\usepackage[ruled,vlined]{algorithm2e}

\begin{document}
\DeclareRobustCommand{\IEEEauthorrefmark}[1]{\smash{\textsuperscript{\footnotesize #1}}}
%
\title{Deep diffusion-based forecasting of COVID-19 by incorporating network-level mobility information}

%
    \author{
        Padmaksha Roy\IEEEauthorrefmark{1},
        Shailik Sarkar\IEEEauthorrefmark{2},
        Subhodip Biswas\IEEEauthorrefmark{2},
        Fanglan Chen\IEEEauthorrefmark{2}, 
        Zhiqian Chen\IEEEauthorrefmark{3}, 
        \\
        Naren Ramakrishnan\IEEEauthorrefmark{4} and
        Chang-Tien Lu\IEEEauthorrefmark{4}
        \\
        \\
     \IEEEauthorblockA{
     \IEEEauthorrefmark{1}Department of Electrical and Computer Engineering, Virginia Tech, Greater Washington DC area, USA\\ E-mail: padmaksha@vt.edu}
    
    \IEEEauthorblockA{
    \IEEEauthorrefmark{2}Department of Computer Science, Virginia Tech, Greater Washington DC area, USA\\ E-mail(s):shailik@vt.edu, subhodip@cs.vt.edu, fanglanc@vt.edu}

    \IEEEauthorblockA{
    \IEEEauthorrefmark{3}Department of Computer Science and Engineering, Mississippi State University, Starkville, Mississippi, USA\\
    Email: zchen@cse.msstate.edu}
    \IEEEauthorblockA{
    \IEEEauthorrefmark{4}Sanghani Center for Artificial Intelligence and Data Analytics, Virginia Tech, Greater Washington DC area, USA\\ E-mail(s): naren@cs.vt.edu, ctlu@cs.vt.edu}
    }


\newcommand{\algo}{{\textsf{ARM3Dnet}}}
\newcommand{\etal}{{\textit{et al}.~}}

\maketitle

\pagestyle{plain}

\begin{abstract}

Modeling the spatiotemporal nature of the spread of infectious diseases can provide useful intuition in understanding the time-varying aspect of the disease spread and the underlying complex spatial dependency observed in people's mobility patterns. Besides, the county level multiple related time series information can be leveraged to make a forecast on an individual time series. Adding to this challenge is the fact that real-time data often deviates from the unimodal Gaussian distribution assumption and may show some complex mixed patterns. 
Motivated by this, we develop a deep learning-based time-series model for probabilistic forecasting called Auto-regressive Mixed Density Dynamic Diffusion Network~(\algo), which considers both people's mobility and disease spread as a diffusion process on a dynamic directed graph. The Gaussian Mixture Model layer is implemented to consider the multimodal nature of the real-time data while learning from multiple related time series. We show that our model, when trained with the best combination of dynamic covariate features and mixture components, can outperform both traditional statistical and deep learning models in forecasting the number of Covid-19 deaths and cases at the county level in the United States.
\end{abstract}

%

\section{Introduction}
The worldwide outbreak of the COVID-19 pandemic in 2020 has restricted people mobility both in micro and macro levels in an unprecedented manner. Although the origin of the disease still remains unclear but the unpredictable nature of the spread of the disease depending on various factors needs further investigation. A reliable disease forecasting of the outbreak can aid not only in preparing for adopting containment measures but also in identifying areas with a high risk of spread for better resource allocation~\cite{nikolopoulos2021forecasting}.

The scientific community has tackled this problem with different types of disease spread models. While in the beginning, most of the efforts were focused on traditional mathematical modeling and simulation~\cite{cooper2020sir}, the application of data-driven methods increased as more data became widely available.  While the majority of the works focused on forecasting the exact number of active cases~\cite{roy2020spatial,Melin2020-kc}, some focused on predicting the number of newly infected cases, rate of infection, and a number of incoming cases from different geographical regions~\cite{kim2020hi}. 
Our work proposes a data-driven method to forecast the exact number of positive cases and deaths at a multitude of different geographically smaller regions using spatiotemporal information of multiple correlated time series and people mobility information across those regions.

 Most of the work on COVID-19 forecasting has so far exploited the temporal nature of the data and to the best of our knowledge have not considered both the spatial dependency among different region and the correlation among different time series across counties together to achieve better forecasting results. Adding to this is the uncertainty aspect of such forecasting in presence of various static and dynamic features. 
    
    \begin{figure}[htpt!]
      \centering
      \includegraphics[width=0.35\textwidth]{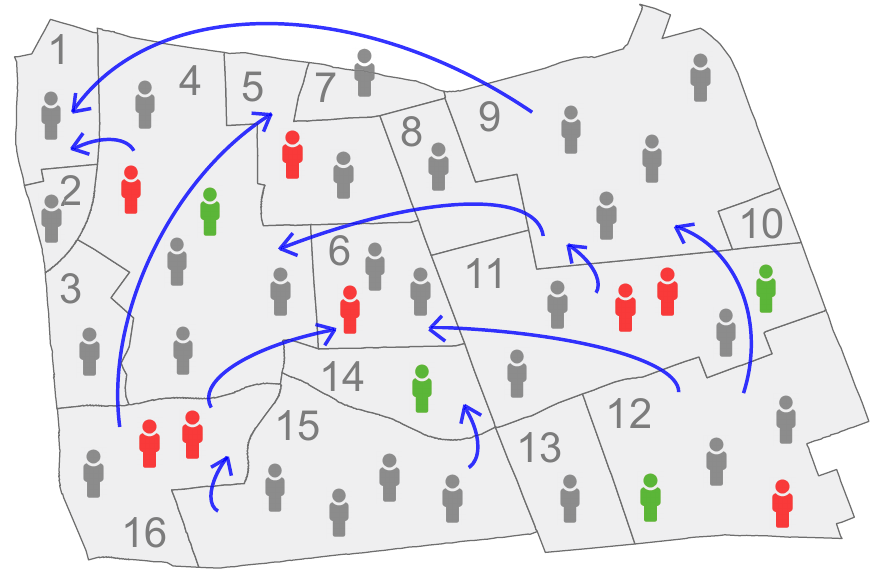}
      \caption{A simulation depicting our assumption that infected people's mobility(depicted in red in the diagram) within and across counties is correlated with the disease spread due to contagious nature of the disease.} 
      \label{fig:concept}
    \end{figure}

In the early part of the pandemic, several factors like race and gender, \cite{rossen2020excess}, underlying factors$-$socioeconomic status~\cite{khalatbari2020importance}, lock-down policy enforcement~\cite{das2020prediction}, social distancing measures~\cite{thunstrom2020benefits}, change in community mobility in different regions$-$that have proven to contribute more towards the nature of the spread~\cite{kapoor2020examining}. Our work mainly focuses on the observation how people's mobility at granular levels in between neighboring counties can have a direct correlation with the spread of the disease. 
 The infectious and contagious nature of COVID-19 makes us hypothesize that understanding the population diffusion process can be significant in understanding the spread of disease. 
 
 In our work, we model the dynamics of people mobility as a diffusion process and use the diffusion convolution operation to capture the strong spatial dependency. We further propose the ~\algo~that integrates the diffusion convolution operation and an auto-regressive recurrent mixed density network and use sampling techniques to capture the true statistical nature of the data.  We formulate our problem on Graph Neural Networks (GNNs) that can be utilized to get more accurate forecasts as they integrate graph structure with node attributes~\cite{scarselli2008graph,gcn}. The dynamic data collected by us, can be leveraged for implementing a graph structure with every node representing a county in the United States and edges denoting the people's mobility between the counties. 
 

The proposed model is a global model on multiple sequences of time series data, each corresponding to a county in the US~\cite{bandara2017forecasting}. 
A challenge with forecasting any infectious disease is that, with the ROIs (Region of Interest) becoming smaller, forecasting across different geographical regions becomes less consistent, making it poorer for generalization. Again, with an increased number of time series sequences, both the magnitudes and the distribution of the magnitude differ widely~\cite{salinas2020deepar}. In fact, the majority of the existing works focus on forecasting of COVID-19 infection rate at a state level or country level~\cite{shahid2020predictions,chimmula2020time} and very few works have focused on proposing a global model that leverages the spatial as well as temporal dependency between geographical locations, and the joint training of multiple correlated time series data.


Based on the above-mentioned observations, we develop our~\algo~model in an attempt to address them. The contributions of this work can be summarized as follows:  

\begin{itemize}
    \item \textbf{We show that there is a direct correlation of inter and within county people's mobility with the spread of the disease and associated deaths and infections at granular level.}
    We model the spatial dependency of people's mobility as a diffusion process on a directed graph formed with a certain number of counties. The diffusion convolution intuitively captures both spatial and temporal dependencies when fed as time-dependent co-variate to a sequence-sequence learning framework. We experimented with our county-level graph network adding different amounts of sparsity and edge weight-ages and our results are reported in the later sections.

    \item \textbf{We try to prove that by considering different county-level time series which can be correlated and via sharing this information, one can actually improve the forecasting accuracy and the uncertainty.}
    In order to achieve it, we created an auto-regressive model that can jointly train multiple time series together and as a result at every instant the model learns from the time series information of neighboring counties. Our findings are reported in the subsequent sections.

    \item \textbf{We propose a model which can capture the multi-modal nature of the COVID real-time data that shows multiple peaks over time.}
    We introduced the Gaussian Mixture Model on top of the auto-regressive model to model the real time multi-modal nature of the COVID data as reported by the CDC and found improvements both in terms of accuracy and uncertainty. This ultimately strengthens our assumption that real time COVID data was indeed having multiple peaks and cannot be modelled as a single mode Gaussian distribution.

    \end{itemize}

\section{Background}
\label{sec:background}

Influenza-like illness (ILI) has been a subject of epidemiological research for several years. Flu forecasting, for example, has used everything from a mechanistic model that leverages epidemiological insights~\cite{shaman2011absolute,zhang2017forecasting} to predefined statistical models for disease modeling based on just historical data~\cite{chakraborty2014forecasting,brooks2015flexible}. Forecasting models using ARIMA and Prophet time series model~\cite{9225319} have also been proposed. But the mechanistic models require a lot of calibration, and they usually struggle to generalize well and accurately fit the data. 

 Cooper \etal\cite{cooper2020sir} studied the effectiveness of the modeling approach on the pandemic due to the spreading of the novel COVID-19 disease. They developed a susceptible-infected-removed (SIR) model that provides a theoretical framework to investigate its spread within a community. On the other hand, Liu \etal \cite{liu2020forecasting} combined ``COVID-19 case data with mobility data to estimate a modified susceptible-infected-recovered (SIR) model in the United States''. They observed that in contrast to a standard SIR model, the incidence of COVID-19 spread is concave if people have inter-related social networks. Recently, Goel \etal\cite{sirASONAM} proposed a mobility-based SIR model for epidemics, which especially takes into account the population distribution and connectivity of different geographic locations across the globe. Although these methods are based on traditional disease simulation models, they often fail to fit the real-world data successfully.
 
 
 Bhouri \etal \cite{Bhouri2020.09.20.20198432} used Google's mobility data to use as features for their proposed DL-based model for doing county-wise forecasting of positive cases for over 200 counties. However, they do not address the correlation among different time series or the spatial dependency among different counties. Works like Adhikari \etal\cite{adhikari2019epideep} addressed these issues and proposed a data-driven DL-based model that leveraged the seasonal similarities of historical data for flu forecasting tasks and introduced model interpretability. Although COVID-19 had a similar spreading pattern to other ILIs, the complete lack of historical seasonal data adds to the challenge of accurate forecasting. The data-driven methods suffer from exploiting seasonal similarities and providing enough interpretability and fail to provide an uncertainty estimation of the forecasting over a long period of time.


 The Deep-AR model \cite{salinas2020deepar} has successfully modeled multiple related time series data for traffic and electricity forecasting of households.  Mixed Density Networks have been used to model complex data distributions that deviate from single-mode Gaussian patterns to an arbitrary conditional density function.\cite{bishop1994mixture,mclachlan1988mixture}. In many of such cases, the time series of the neighboring nodes are correlated and through joint training of multiple correlated time series, a substantial amount of data on past behavior of similar, related time series can be leveraged for making a forecast of an individual time series. Again in the spatial domain, if one considers the mobility flow between neighboring counties as a diffusion process that can capture the stochastic nature of the population dynamics.\cite{li2017diffusion} 

Motivated by the above, we propose the \algo~algorithm to model the spread of COVID-19 at multiple geographical regions simultaneously while also modeling the correlation among different time series sequence and the spatial dependency among these regions. 
 

  \begin{figure*}[htpb!]
  \centering
  \includegraphics[width=0.975\textwidth]{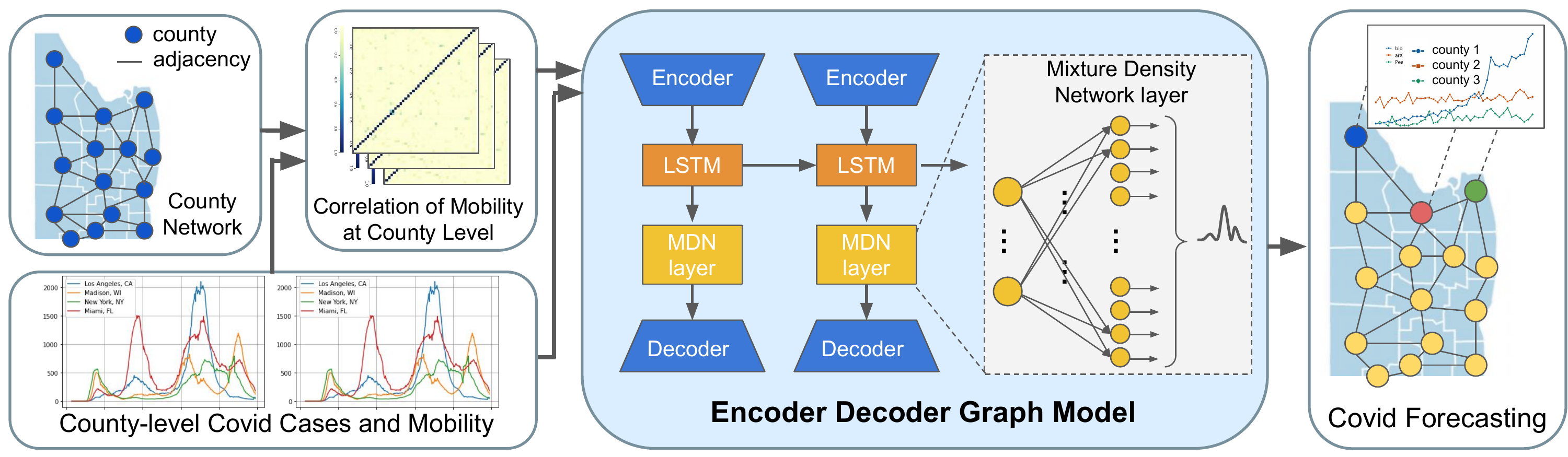}
  \caption{An outline of the \algo~forecasting framework. The input to the network are the graph convoluted covariates, the target value of the previous step and the previous network output. The encoder network is followed by a MDN layer to capture the multimodal data. The output of the network is used to compute the parameters of the likelihood function.}
  \label{fig:model}
\end{figure*}

\section{Methodology}
\label{sec:method}
In this section, we outline the problem of forecasting COVID-19 disease at a county level in the US. This is followed by a step by step description of our model.

\subsection{Problem Formulation}
Graphs are used to represent a wide variety of complex systems with interacting entities. 
We create a county level graph network where the relationship between the county locations is represented with a directed graph $\mathcal{G}$, where $\mathcal{G} = \left(\mathcal{V},\mathcal{E},\mathcal{A}\right)$ is a set of $N$ nodes representing each county in the US, $\mathcal{V}$ is a set of nodes $|\mathcal{V}| = N$ and $\mathcal{E}$ is a set of directed edges representing the connection between the nodes. $\mathcal{A} \in {\mathbb{R}}^{N \times N}$ is the weighted adjacency matrix representing the strength of connectivity between the nodes. The edges of the graph are connected on the basis of aggregated visits of people from an origin to a destination county on a  particular day as gathered from the SafeGraph Community. Our model is based on two fundamental assumptions, firstly, the diffusion pattern across neighboring nodes has a fundamental statistical property where, after many time steps, such Markovian process converges to a stationary distribution $\rho \in \mathcal{R}^{N \times N}$, whose $i$\textsuperscript{th} row, $\rho_i \in \mathcal{R}^N$, 
 represents the likelihood of diffusion from node $v_i \in \mathcal{V}$ and therefore proximity with respect to node $v_i$. Secondly, due to the extremely contagious nature of the disease, the people's mobility at the county level and the population diffusion in between the neighboring counties is bound to have some correlation with the spread of the disease in general. Therefore, given the historical covariates, like aggregated number of infection cases $X_{i,t}$, the target value at the previous time step $Z_{i,t-1}$, and the previous network output $h_{i,t-1}$, the goal of the network is to learn the following function:

In our model, the time covariate feature of the AR-LSTM is replaced by a diffusion convolution operation which is actually a graph convolution between county level people mobility information and the adjacency matrix which carries the spatial information. The overall architecture of our~\algo~model is shown in Figure~\ref{fig:model}. It utilizes a graph-based encoder-decoder architecture that models the spatial correlation by a diffusion convolution operation on the graph and the temporal patterns by an auto-regressive long short-term memory (AR-LSTM)  network. The goal of our study is to create a global model that can learn from this number of related time series together.

\begin{align*}
    h_{i,t} = \mathbf{h}\left(h_{i,t-1},Z_{i,t-1},X_{i,t},\Theta\right)  
\end{align*}

    \begin{figure}[htpt!]
      \centering
      \includegraphics[width=0.45\textwidth]{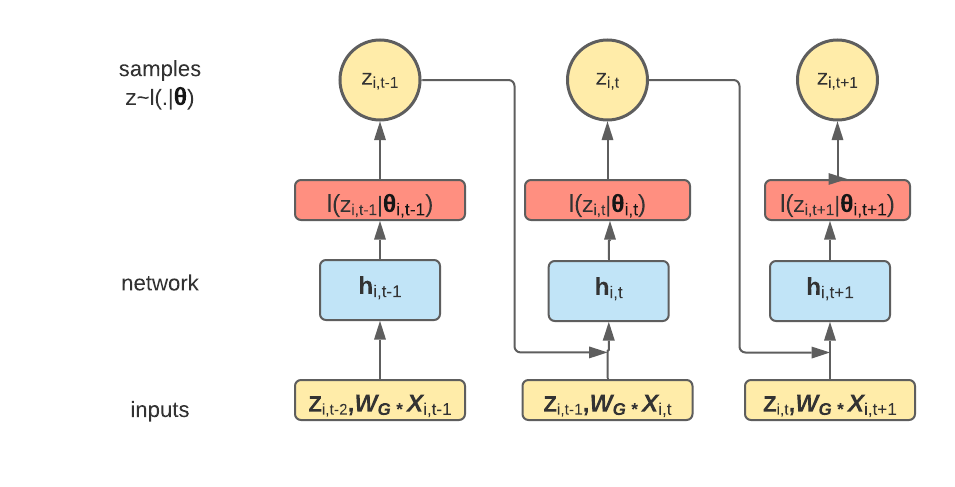}
      \caption{A block diagram of the auto-regressive LSTM encoder-decoder network to jointly model the correlated time series information.} 
      \label{fig:concept}
    \end{figure}


\subsection{Capturing the spatio-temporal correlation}
The fundamental idea of a convolution layer in a neural network is to extract localized features from the input data by convolving with filters of various dimensions. Similarly, the core idea of a graph convolution layer is to extract the localized features from input data which is being represented as a graph structure. Therefore, the product of a neighborhood matrix $\mathcal{A}$, input data $\mathcal{X}_t$, and a trainable weight matrix $\mathcal{W}$, is considered as a graph convolution operation to extract features from a single hop neighborhood. 
\begin{equation}
    \mathcal{W_G} * \mathcal{X} = \mathcal{W}(\mathcal{D}^{-1} \mathcal{A}) \mathcal{X}
    \label{eq:eq1}
 \end{equation}

Here, $\mathcal{D}$ is the degree matrix, $\mathcal{D}^{-1}\mathcal{A}$ is a transition matrix of the diffusion process, and $\mathcal{W}$ is a learnable filter for the diffusion process. Since the people's mobility across the counties are dynamic in nature, therefore our ~\algo~ model uses a dynamic weighted adjacency matrix. We keep the adjacency matrix as dynamic by computing the pearson correlation of the aggregated visits among the counties on daily basis. Therefore, the elements in the adjacency matrix $\mathcal{\Tilde{A}}$ at time ${T}$ are given by,
\begin{align*}
    \mathcal{A}_{i,j} = \rho_{{x}_{i},{x}_{j}} = \frac{cov({x}_{i},{x}_{j})}{\sigma_{{x}_{i}},\sigma_{{x_j}}}
\end{align*}
where, $\mathcal{A}_{i,j}$ represents the edge weight between the nodes $i$ and $j$; ${x}_i$ and ${x}_j$ denotes the aggregated visits in between nodes $i$ and $j$. The Pearson correlation co-efficient between $x_i$ and $x_j$ is represented by $\rho$; $cov$ is the covariance, $\sigma_{x_i}$ and $\sigma_{x_j}$ represents the standard deviation of the nodes $x_i$ and $x_j$, respectively.

We found that a sparse and weighted adjacency matrix truly captures the correlation between the people diffusion, disease spread and the number of death forecasts. In total, our graph has 300 nodes and we have connected an edge between two counties whenever the aggregated visits people made between them crossed a threshold of 200 on a particular day. The graph convoluted output is fed to an auto-regressive recurrent encoder and decoder model.

\subsection{Modeling the likelihood function}
Given, the value of time series $\textit {i}$ at time $\textit{t}$ by ${Z_{i,t}}$, our goal is to model the conditional distribution:
\begin{align*}
    \mathcal{P}\left (\mathcal{Z}_{i,t_0:T} | \mathcal{Z}_{i,1:t_0-1}, \mathcal{W_G*X}_{i,1:T}\right)
\end{align*}
where, $Z_{i,t_0:T}$ and $\mathcal{Z}_{i,1:t_0-1}$ represents the future and the past time series, respectively.
Here, the dynamic covariates $\mathcal{W_G*X}$ are the graph convolution between aggregated people's mobility at the county level on daily basis and the county level adjacency matrix. During training, the data for the entire conditioning range  $ \left[1,t_{_0}-1 \right ]$ and
prediction range $\left [t_0, T\right] $ are observed but during the forecasting period $\mathcal{Z}_{i,t}$ is available only for the conditioning range.
We assume that the model distribution $Q_\theta(Z_{i,t_0 : T} | Z_{i,1:t_0 -1}, W_G*X_{i,1:T})$ consists of a product of the likelihood factors:
\begin{equation}
 \begin{split}
Q_\Theta(\mathcal{Z}_{i,t_0 : T} | \mathcal{Z}_{i,1:t_0 -1},  \mathcal{W_G*X}_{i,1:T}) =
 &
 \\
 \prod_{t=t_0}^T Q_\Theta (\mathcal{Z}_{i,t}|\mathcal{Z}_{i,1:t-1}, \mathcal{W_G* X}_{i,1:T})
 =  &
\prod_{t=t_0}^T \ell(\mathcal{Z}_{i,t}|\theta(h_{i,t},\Theta))
 \\
 \end{split}
 \label{eq:eq2}
\end{equation}
which is parameterized by the output $h_{i,t}$ of an autoregressive recurrent network.

\begin{align*}
    h_{i,t} = \mathbf{h} \left(h_{i,t-1},\mathcal{Z}_{i,t-1}, \mathcal{W}_G * \mathcal{X}_{i,t},\Theta \right)
\end{align*}

The network output $h_{i,t},\mathcal{Z}_{i,t-1}$ is used to compute the parameters $\theta_{i,t}$ of the likelihood $l(\mathcal{Z}|\theta)$ and that is used for training the model parameters.
The likelihood $\ell\left(\mathcal{Z}_{i,t}|\theta\left(h_{i,t}\right)\right)$ is a fixed distribution whose parameters are defined by the function $\theta\left( h_{i,t}, \Theta \right)$ of the network output $h_{i,t}$.
Given the model parameters $\Theta$, joint samples are obtained through ancestral sampling as
\begin {equation}
\tilde {\mathcal{Z}}_{i,t_0:T}\;\sim\; Q_\theta(\mathcal{Z}_{i,t_0 : T} | \mathcal{Z}_{i,1:t_0 -1},  \mathcal{W_G*X}_{i,1:T}).
\label{eq:eq3}
\end {equation}

For $t = t_0, t_0 + 1, \ldots, T$, the sampling is performed as $\tilde{ \mathcal{Z}}_{i,t}$ $\sim$ $\ell \left(.|\theta(h_{i,t}, \Theta)  \right)$.
Samples obtained from the model can be used to calculate the quantities of interest from the posterior distribution for a given time range in the future.

In this context, the likelihood function $l\left(\mathcal{Z}|\theta\right)$ must be carefully chosen to match the statistical properties of the data. The network predicts the parameters mean and variance of the probability distribution for the next time point. In our case, we have chosen the Gaussian likelihood for predicting the death count data which is parameterized using its mean and standard deviation as given below:
\begin{align*}
    l_G\left(\mathcal{Z}|\mu,\sigma\right) & = \frac{1}{\sqrt{2 \pi \sigma^2}}  \exp{\left(-\frac{{(\mathcal{Z}-\mu)}^2}{2\sigma^2} \right)}
\\
    \mu\left(h_{i,t}\right) & = W_\mu^T h_{i,t} + b_\mu
\\
    \sigma\left(h_{i,t}\right) & = \log \left(1 + \exp\left({W_\sigma}^T h_{i,t} + b_\sigma \right)\right)
\end{align*}
The standard deviation is often followed by a softplus activation function in order to ensure the value is positive always.

\subsection{Mixture Density Layer to model the multimodal data} 

If we assume the conditional distribution of the target data to be Gaussian, we can estimate the parameters using the maximum likelihood. But it gives the idea of replacing the conditional distribution of the target vector with a mixture model that is flexible to completely model a general distribution function. Thus, we model the probability density of the target data as a linear combination of the Kernel functions.
\begin{align*}
    \mathcal{P}(\mathcal{Z}|\mathcal{X}) = \sum_{i=1}^m \alpha_i(x) \cdot p_i(\mathcal{Z}|\mathcal{X})
\end{align*}
where, $m$ is the number of components in the mixture and $\alpha_i(x)$ are the mixing coefficients. When the Kernel functions are assumed to be Gaussian in nature, the equations are of the form:
\begin{align*}
    p_i(\mathcal{Z}|\mathcal{X}) = \frac{1}{(2\pi)^{c/2} \sigma_i(x)^c} \exp \left(-\frac{||\mathcal{Z}-\mu_i(x)||^2}{2\sigma_i(x)^2}\right)
\end{align*}
where $\mu_i$ represents the centre of the $i^{th}$ kernel.

Given this information, a Gaussian Mixture Model (GMM) with this simplified kernel can approximate any given density function to arbitrary accuracy, providing the mixing coefficients and the Gaussian parameters are correctly chosen~\cite{bishop1994mixture}. Although the output of the Auto-regressive LSTM model minimizes a softmax loss function, these losses are not suitable for modeling the variation in the data over a long period of time. Thus a single Gaussian distribution often fails to model the multimodal nature of the real world data. Therefore, our main hypothesis for modeling the output distribution of the parameters is performing the regression on a multimodal data. Our model combines a traditional AR-LSTM model with a mixture density model that gives a general framework for modeling conditional density functions. Therefore, the conditional distribution for $K$ Gaussian mixtures can be written as:
\begin{align*}
    \mathcal{P}\left( \mathcal{Z}_{i,t}| h_{i,t} \right)  \sim p_1 \mathcal{N}\left(\mu_1(x), \sigma_1^2(x) \right) + p_2 \mathcal{N}\left(\mu_2(x), \sigma_2^2(x) \right) \\ 
    \ldots + p_k \mathcal{N}\left(\mu_k(x), \sigma_k^2(x) \right) +
    \ldots + p_K \mathcal{N}\left(\mu_K(x), \sigma_K^2(x) \right)
\end{align*}
where, $p_k$, $\mu_k$ and $\sigma_k$ are the probability, mean and standard deviation respectively for the $k$th Gaussian component.
The parameters of the GMM are computed from the $h_{i,t}$ output of the recurrent layer as follows:
\begin{align*}
    p_k = \frac{exp\left(\theta_k^{(p)}h \right)}{\sum_{l=1}^K \theta_l^{(p)}h }
\\
    \sigma_k = \exp\left(\theta_k^{(\sigma)} h\right)
\\
    \mu_k = \exp \left(\theta_k^{(\sigma)} h\right)
\end{align*}
where, $\theta_k^{(p)}$, $\theta_k^{(\sigma)}$ and $\theta_k^{(\mu)}$ represents the learned parameter of the output layer corrosponding to the mixture probability, variance and mean respectively. 

\subsection{The \algo~algorithm and the loss function}
Given the time series ${\mathcal{Z}}_{i,1:T},\;{i=1.,..N}$ and associated graph convoluted covariates $\mathcal{G{C}}_{i,1:T}$, where the time range is chosen in such a way that $\mathcal{Z}_{i,t}$ in the prediction range is known, the parameters $\Theta$ of the model can be learnt by maximising the log-likelihood function.

\begin{equation}
    \mathcal{L}= \sum_{i=1}^N \sum_{t=t_0}^T \log \left(\sum_{k=0}^m \alpha_k \left(\theta\right) \cdot \ell{ \mathcal(Z_{i,t}|\theta(h_{i,t})} \right)
    \label{eq:eq4}
\end{equation}
where, $h_{i,t}$ is a deterministic function of the input and can be optimized directly via the stochastic gradient descent algorithm during backpropagation by computing the gradients with respect to $\Theta$. The pseudocode of \algo~is provided in Algorithm~\ref{algo}. 

\begin{algorithm}[htp!]
\SetAlgoLined
\textbf{Input:  $\mathcal{X} \in \mathcal{R}^{N \times P},A \in \mathcal{R}^{ N \times N}$ }\;
\textbf{Parameters: $\mathcal{W}_{g_{c_{1}}},..., \mathcal{W}_{g_{c_{n}}}, \mathcal{W}, b$}\;
\textbf{Initialize: ${h_0 = 0, C_0 = 0 \in \mathcal{R}^N }$}\;
 \For{$t =1$ to T}
 {
     $\mathcal{\tilde{A}} = \mathcal{D}^{-1}\mathcal{A}$\; 
     $\mathcal{G{C}}_t =\left ( \mathcal{W}_{g{c}} \circ \mathcal{\tilde{A}} \right) \mathcal{X}_t$\; 
      
     $h_{i,t} = \mathbf{h}(h_{i,t-1},\mathcal{Z}_{i,t-1},\mathcal{G{C}}_{i,t},\Theta)$ \;
     $\tilde {\mathcal{Z}}_{i,t}\;\sim\; Q_\Theta(\mathcal{Z}_{i,t} | \mathcal{Z}_{i,t -1},  \mathcal{G{C}}_{i,t}).$ 
     
    \emph {calculate $\sigma$} \;
 }
 return $\tilde {\mathcal{Z}}_{i,t_0:T}$\
 \caption{\algo~Model}
 \label{algo}
\end{algorithm}

For each of the county level time series data of covid, we create multiple training instances by selecting windows with different starting point from the original time series. Here, during training the total length of the time series $T$ and the conditioning and prediction ranges are kept fixed for all training examples. We ensure that the entire prediction range is always covered by the ground truth data.
During training, the values of ${\mathcal{Z}}_{i,t}$ are known in the prediction range and can be used to compute the model parameters $h_{i,t}$.
However, during prediction values of ${\mathcal{Z}}_{i,t}$ are unknown in the for the time period $t\geq t_0$ and therefore a single sample
${\tilde{\mathcal{Z}}}_{i,t} \sim \ell \left (. |\theta(h_{i,t})  \right)$ from the model distribution is used to compute $h_{i,t}$. 
In a multivariate normal distribution, the standard approach is to maximise the log-likelihood function.

\section{Experimentation}
\label{sec:experiment}
In this section, we discuss the datasets, experimental settings, performance metrics, the baseline methods, and their parametric settings, and the comparative results. Our implementation of the proposed model will be available at \url{https://github.com/padmaksha18/ARM3Dnet}

\subsection{Datasets}
\label{sec:dataset}
In this study, we focus on county-level COVID-19 data and point-of-interest (POI) based Mobility data developed by SafeGraph.  Here, we describe our data sources, methods of collecting them as well as two of our curated datasets (ML-Dataset and Covariate Feature Dataset) designed for its use in mobility-based future work on forecasting.

\paragraph{\textbf{COVID-19 data}} 
We collected the data from the public repository maintained by JHU  and NYT. We used cumulative cases and deaths for each date and each county, from May to December of 2020.


\paragraph{\textbf{SafeGraph POI based Mobility Data}} We used the data from SafeGraph data (https://www.safegraph.com)\ to build a human mobility network. SafeGraph uses device location to build a dataset that details the number of visits to a POI from all other Census Block Groups (CBGs) on every single day based on a number of devices. 

Using the POI based mobility data, the final derived dataset had attributes like \textsf{Date}, \textsf{Origin County FIPS}, \textsf{Destination County FIPS}, \textsf{Mean Distance traveled by a device} (corresponding to Origin County), \textsf{Aggregated Visits} (from origin to destination) and \textsf{Device Count}.
\textsf{Aggregated Visits} is used as the metric of inter county mobility flow whereas \textsf{Mean Distance} is considered intra-county mobility feature.

\paragraph{\textbf{Curated Mobility Dataset}} We create two different datasets using the aforementioned mobility features for different sets of models.

\textbf{ML-Dataset:} This dataset is used as mobility features for DeepCOVID-19, one of our baselines. The features are a combination of intra-county and inter-county mobility. The intra-county mobility feature is "mean distance traveled by a device" from SafeGraph. 
We derive 'inflow' as inter-county mobility corresponding to each county. We define inflow at time $t$ and county $i$, as $\sum_{j}^{}X_{j,i}*C_j$( where $X_{j,i}$is the aggregated visit where origin county is $j$ and destination county is $i$; $C_j$ is the number of cumulative cases for county $j$).

\textbf{Mobility-convoluted covariate features:} The covariates in our case are time-dependent. It is based on the assumption that both inter and intra-county people mobility plays a key role in disease propagation (number of cases, here) through the county mobility network where each node is a county. 
The covariate dynamic time series information, when graph convoluted with the county level adjacency matrix formed on the basis of the weights given to the number of people traveling between them, will provide additional information to the model during prediction. The covariates are standardized to follow zero mean and unit variance and thus we create for each day a covariate dataset which is used by~\algo~and DC-RNN.

\subsection{ Experimental Settings:} We used COVID-19 cumulative case and death data from May to December and experiment with different length of training data to do prediction for the next 7 days. We also changed the sparsity of the adjacency matrix by incorporating more counties in the graph and observed how our proposed model gave a better performance with an increase in the number of nodes.  For the mobility features, we use inter and cross county mobility from publicly available data from SafeGraph. 

\subsection{Baseline Models}
We use the following baseline methods for comparison.
\begin{itemize}
    \item \textbf{SARIMAX}: This is a statistical auto regressive model for time series prediction. We use Augmented Dickey-FUller Test to check if the time series data is stationary and then use grid-search for the selection of parameters. 
    \item \textbf{LSTM}: Traditional RNN model ideal for sequential time series data. We consider a time window of 15 days.
    
    \item \textbf{DeepCovid-19}~\cite{Bhouri2020.09.20.20198432}: This is an existing deep learning-based model for COVID forecasting, that was designed for county level forecasting of positive cases using mobility features.  Similar to LSTM, we use our ML-Dataset as a mobility feature.
    \item \textbf{DCRNN}~\cite{li2018diffusion}: This is a state-of-the-art deep learning framework for spatiotemporal forecasting.  Our mobility-based Covariate feature dataset is used as the input from the graph convolution element of the model. 
    \item \textbf{DeepAR}~\cite{salinas2020deepar}: This is a state-of-the-art time series prediction model that deals with multiple correlated time series. We use the same covariate features for DeepAR as described earlier. 
\end{itemize}
  
  We simulate these baselines using the hyperparameter settings recommended by the literature. For our model, each covariate feature matrix is $\in \mathcal{R}^{T \times 1}$ where each row corresponds to the number of days in the time series data and the columns represent dynamic time series features like graph convoluted mobility features. In some cases, we found that considering the disease time series(number of cases) in the graph convolution operation also helps in improving uncertainty of prediction which indirectly supports the contagious nature of the disease. Under Gaussian likelihood, we use the Adam optimizer with early stopping to train the model for 50 epochs.
 
\subsection{Metric for Evaluation:} 
We compared the accuracy of our forecasting on both numbers of cases and deaths using Normalized Root Mean Squared Error (NRMSE) and Normal Deviation (ND). The scale of the county level time series varies widely and therefore the normalized RMSE was considered. The uncertainty growth over time from the data was measured and compared with the ND metric to gauge the confidence of the predictions over the entire time period. They are calculated as follows.

\begin{align*}
    \text{NRMSE} = \frac{\sqrt{\frac{1}{N(T-t_0)}\sum_{i,t}(\mathcal{Z}_{i,t} - \mathcal{\hat Z}_{i,t})^2}}{\frac{1}{N(T-t_0)}\sum_{i,t}|\mathcal{Z}_{i,t}|},
\end{align*}

\begin{align*}
    \text{ND} = \frac{\sum_{i,t}\left|\mathcal{Z}_{i,t} - {\hat{\mathcal{Z}}}_{i,t}\right|}{\sum_{i,t}|\mathcal{Z}_{i,t}|}.
\end{align*}

    

 \subsection{Results \& Discussions}

The results of our baselines comparison, which are normalized over 300 time series data, clearly shows that our proposed model significantly outperforms traditional deep learning and statistical models and even challenges state of the art models like Deep AR on the basis of ND and NRMSE. The purely statistical models like SARIMAX and the vanilla LSTM model performs worse in comparison to the other deep learning models like DeepCovid-19 and DC-RNN, where the inter and intra county mobility information has been incorporated as a feature during modeling. This justifies the need of the granular level mobility information while formulating such a problem.

\begin{table}[htbp!]
\caption{Predictive results of \algo~and other baseline methods in forecasting the number of cases and deaths due to COVID-19. The values are normalized over 300 time series.}
\centering
\small
\label{tab:results}
        \begin{tabular}{c|cc|cc}
        \toprule
        \multirow{2}{*}{\diagbox{Models}{Metrics}}
       & \multicolumn{2}{c|}{\bf Cases Forecasting} & \multicolumn{2}{c}{\bf Deaths Forecasting} 
       \\
       & NRMSE & ND & NRMSE & ND 
        \\ \hline
        SARIMAX & 0.210 & 0.033 & 0.129 & 0.070 
        \\ 
        Ensembled LSTM & 0.368 & 0.064 & 0.203 & 0.065
        \\
        DC-RNN & 0.323 & 0.046 & 0.096 & 0.033
        \\
        DeepCovid19 & 0.189 & 0.065 & N/A & N/A
        \\
        DeepAR & 0.170 & 0.035 & 0.054 & 0.024 
        \\ 
        \algo & \textbf{0.091} & \textbf{0.024} & \textbf{0.051} & \textbf{0.018} 
        \\ \bottomrule
        \end{tabular}
        \normalsize
\end{table}

\begin{figure}[htpb!]
	\centering
    \begin{subfigure}[b]{0.44\linewidth}
    	\centering
        \includegraphics[keepaspectratio,width=\textwidth]{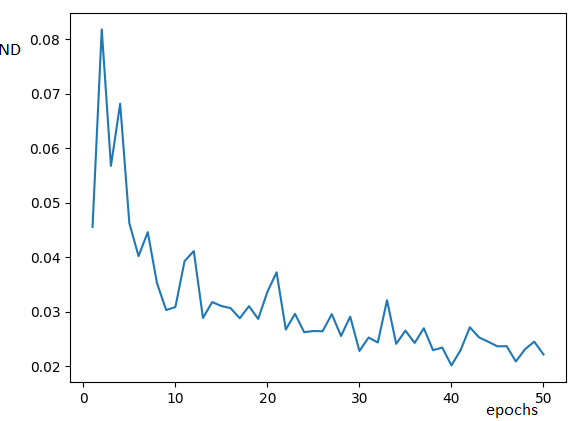}
        \caption{with dynamic covariates}
    \end{subfigure}	
~
    \begin{subfigure}[b]{0.48\linewidth}
        \centering
        \includegraphics[keepaspectratio,width=\textwidth]{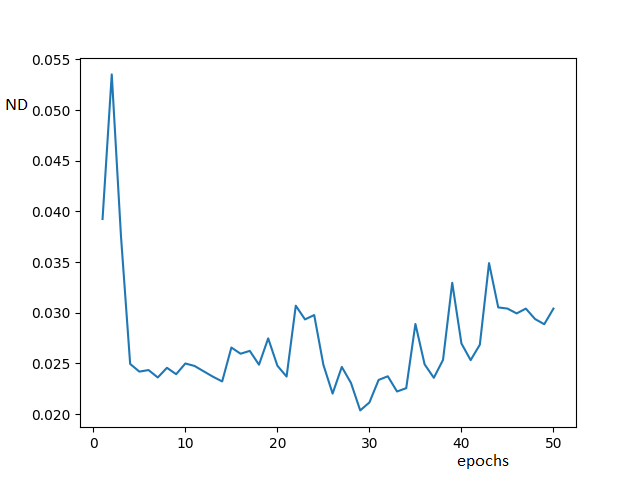}
        \caption{without dynamic covariates}
    \end{subfigure}
    \caption{Illustrating the effects of dynamic covariates in the \algo~performance by plotting ND(dynamically generated at each epoch) vs epochs. We observe that incorporating the dynamic covariates leads to smoother convergence of uncertainty.}
	\label{fig:mobility}
\end{figure}

Our \algo~model outperforms both DeepAR and DeepCovid-19\footnote{DeepCovid-19 is used for predicting cases by using a combination of Deep learning and SEIR model. At any given time they calculate the infected and recovered number of population to get cumulative positive cases, and currently do not provide a method for predicting number of deaths}, proving the fact that, the dynamic graph convoluted covariate features has important correlation with the output time series.  Implementing an autoregressive recurrent neural network and using GMM also proves to improve the forecasting performance as we also manage to outperform DCRNN in terms of accuracy and uncertainty.  Accumulating the mobility diffusion information across nodes and node level people mobility data has helped our model to beat a state of the art Deep AR model on both the metrics as shown in Figure~\ref{fig:mobility} for uncertainty estimation. Since the real time covid data show multiple peaks over the period of a year and is truly multimodal in nature, optimizing the GMM with the correct number of components can accurately model this complex data distribution. Our GMM implementation uses 5 mixture components to accurately model the mixed distribution.

\begin{figure}[htp!]
	\centering
    \begin{subfigure}[b]{\linewidth}
    	\centering
\includegraphics[keepaspectratio,width=\textwidth]{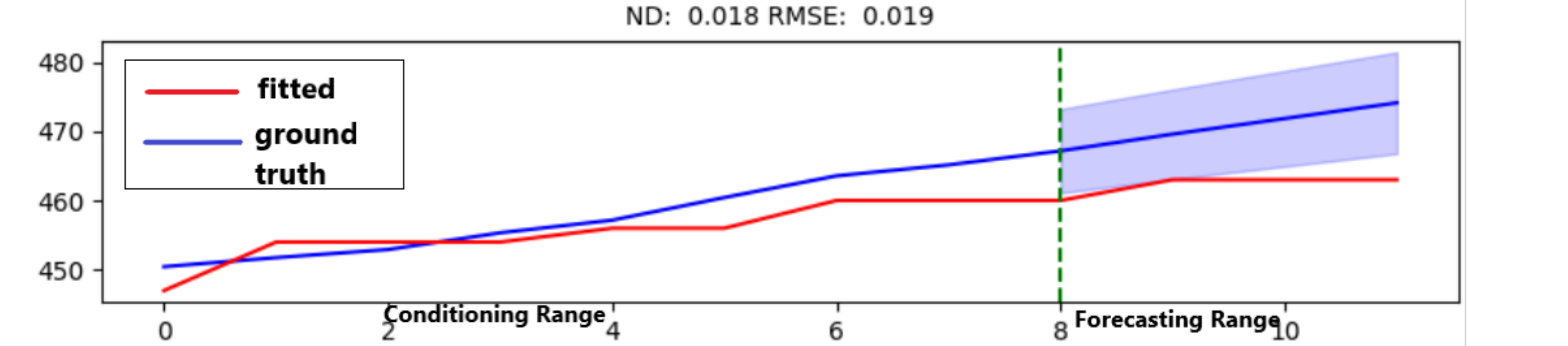}
        \caption{without mobility}
    \end{subfigure}	

    \begin{subfigure}[b]{\linewidth}
        \centering
        \includegraphics[keepaspectratio,width=\textwidth]{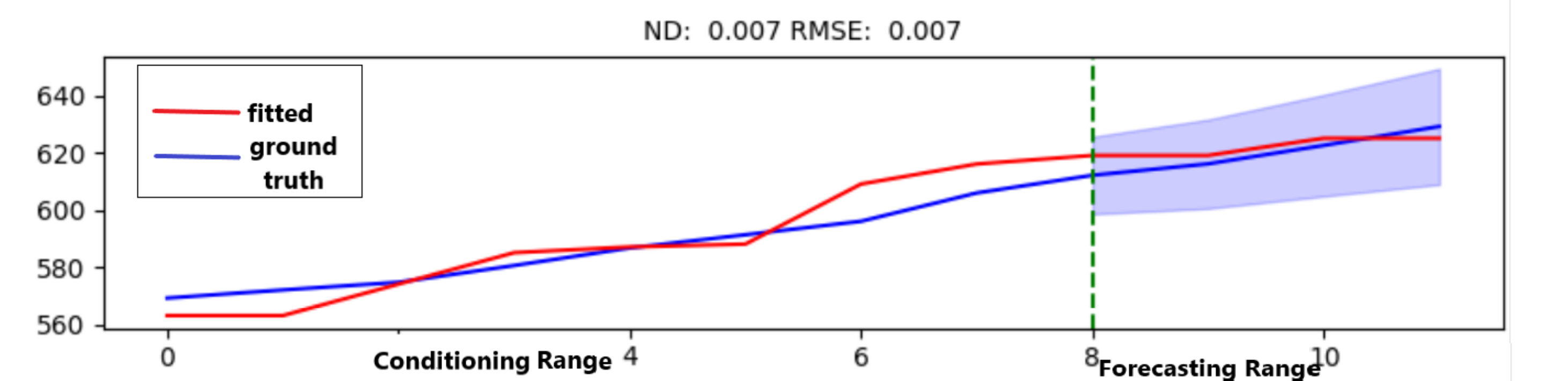}
        \caption{with mobility}
    \end{subfigure}
    \caption{Likelihood estimation for parameters with best ND values (a) with mobility, and (b) without mobility. During stages of training we notice that on adding the mobility information, the model likelihood follows the ground-truth curve more closely in the forecasting range.
    }
	\label{fig:uncertainty}
\end{figure}

We find that the model being aware of the previous information of all the related time series instead of a different model being fitted for each time series independently actually improves both the forecasting accuracy and uncertainty as it can learn the important correlation between all the time series. Thus, sharing information across time series can improve forecast accuracy. The improvement can be seen in the results of our model when compared with an ensemble LSTM model created to train over the different time series independently. The uncertainty growth curve~\Cref{fig:uncertainty} shows that our model likelihood very closely adjusts itself with the ground truth data in the forecasting range, where we have taken 12 samples for every iteration during training when the mobility information is incorporated in the covariates.

\begin{figure}[htpb!]
	\centering
        \includegraphics[trim={0.2cm 0cm 0.4cm 0cm},clip,keepaspectratio,width=0.45\textwidth]{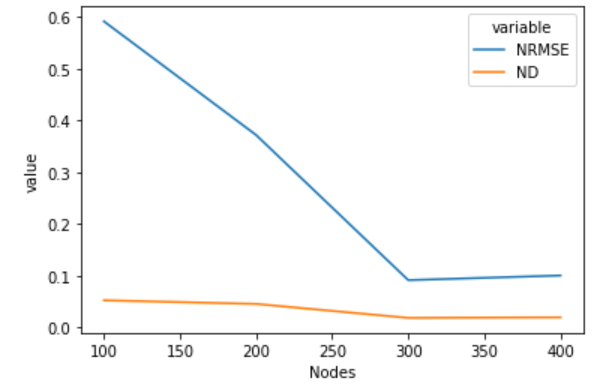}
        \caption{We observe that on adding more number of nodes to the graph network improves performance of our model initially.}
	\label{fig:nodes}
\end{figure}

At its core, our proposed model depends on utilizing the correlation information of the people diffusion between the neighbouring counties and the county level people mobility with the spread of the disease. Therefore we tried to obtain a best threshold for our connected graph network. We found that increasing the sparsity of the adjacency matrix by adding around 300 nodes and assigning more weight-age by connecting only those edges where the aggregated number of visits have crossed 200 on daily basis significantly increases our forecasting results. The results verify our initial assumption that people mobility and the spread of the disease will bear significant correlation during different phases of the pandemic as we can see in \Cref{fig:nodes}.

While training the entire data, we have observed sudden spikes in the loss function during some batches and this can be attributed to the fact that the NYT COVID data has  the aggregated data missing for a consecutive number of days for some counties and are instead replaced by single day count of cases or deaths. We believe that addressing these data-related issues can further improve our model's performance.

\section{Conclusions}
\label{sec:conclusion}
In this paper, we tackle the problem of spatiotemporal forecasting of COVID-19 
at a spatially granular level. We observed that incorporating human mobility network structure and mobility flow information as a covariate feature could empower modern deep learning-based techniques and improve both forecasting accuracy and uncertainty estimation over a long period of time.  The proposed~\algo~framework creates a global model based on multiple related time series data and models the complex spatiotemporal dependency among multiple nodes in the human mobility network. 
In future research, we also want to capture the higher order interactions between static features like socio-economic features and dynamic mobility features and understand its correlation to the disease spread at granular levels.

Future avenues of research may involve using the proposed framework to forecast the spread of other infectious diseases and similar spatiotemporal prediction tasks. 
Also, we believe that  hyperparameter search and optimization~\cite{feurer2019hyperparameter,biswas2020better} of probabilistic disease forecasting models will help to improve the predictive performance, especially in case of the gaussian mixture to choose the optimum set of mixture components.

 \section*{Acknowledgment}
   This research is supported by National Science Foundation grant DGE-1545362. 
   \textbf{Disclaimer:} The views and conclusions contained herein are those of the authors and should not be interpreted as necessarily representing the official policies or endorsements, either expressed or implied, of any institution, NSF, or the U.S. Government.



\bibliographystyle{IEEEtran}
%


\bibliography{literature}

\end{document}